\documentclass[]{spie}  

\usepackage{multirow}
\usepackage{amsmath,amsfonts,amssymb}
\usepackage{graphicx}
\usepackage[colorlinks=true, allcolors=blue]{hyperref}
\title{Deep learning-based auto-segmentation of paraganglioma for growth monitoring}

\author[a,b]{E.M.C. Sijben}
\author[b]{J.C. Jansen}
\author[c]{M. de Ridder}
\author[a]{P.A.N. Bosman}
\author[b]{T. Alderliesten}
\affil[a]{Evolutionary Intelligence Group, Centrum Wiskunde \& Informatica (CWI), P.O. Box 94079, 1090 GB Amsterdam, The Netherlands}
\affil[b]{Dept. of Radiation Oncology, Leiden University Medical Center (LUMC), P.O. Box 9600, 2300 RC Leiden, The Netherlands}
\affil[c]{Dept. of Radiation Oncology, University Medical Centre Utrecht (UMCU), P.O. Box 85500, 3508 GA Utrecht, the Netherlands}

\authorinfo{Further author information:send correspondence to evi.sijben@cwi.nl and/or t.alderliesten@lumc.nl.}

\begin{document} 
\maketitle
\begin{abstract}
\noindent
Volume measurement of a paraganglioma (a rare neuroendocrine tumor that typically forms along major blood vessels and nerve pathways in the head and neck region) is crucial for monitoring and modeling tumor growth in the long term. However, in clinical practice, using available tools to do these measurements is time-consuming and suffers from tumor-shape assumptions and observer-to-observer variation. Growth modeling could play a significant role in solving a decades-old dilemma (stemming from uncertainty regarding how the tumor will develop over time). By giving paraganglioma patients treatment, severe symptoms can be prevented. However, treating patients who do not actually need it, comes at the cost of unnecessary possible side effects and complications. Improved measurement techniques could enable growth model studies with a large amount of tumor volume data, possibly giving valuable insights into how these tumors develop over time. Therefore, we propose an automated tumor volume measurement method based on a deep learning segmentation model using no-new-UNnet (nnUNet). We assess the performance of the model based on visual inspection by a senior otorhinolaryngologist and several quantitative metrics by comparing model outputs with manual delineations, including a comparison with variation in manual delineation by multiple observers. Our findings indicate that the automatic method performs (at least) equal to manual delineation. Finally, using the created model, and a linking procedure that we propose to track the tumor over time, we show how additional volume measurements affect the fit of known growth functions.
\end{abstract}


\section{INTRODUCTION }

Paragangliomas in the head and neck region are rare, often benign, and slow-growing tumors. The locations in which they can appear are categorized as carotid, vagal, jugulotympanic, or a combination of these. If a tumor grows, the risk of adverse symptoms increases. Therefore, tumor growth is an important factor in determining whether treatment is advisable in the short-term~\cite{pellitteri2004paragangliomas,heesterman2019mathematical}. To this end, accurate measurement of the tumor size at multiple follow-up moments on Magnetic Resonance Imaging (MRI) scans is crucial. To measure tumor size, the linear dimensions method is commonly used, supported by clinically available software. With this method an observer measures the biggest diameter of the tumor in three orthogonal dimensions. This technique is much faster than fully delineating the tumor by hand. However, an assumption needs to be made about the tumor's shape in order for the linear dimensions method to be representative of the total volume. In particular, paragangliomas are often assumed to be ellipsoidal. This assumption inevitably leads to estimation errors. Generally, manual volumetric delineation gives more consistent measurements than the linear dimensions method since the volumetric delineation is more precise and thus small variations in measurement have a smaller effect on the measured volume~\cite{heesterman2016measurement}. However, all human 
 measurements, including manual volumetric delineation, suffer, at least to some degree, from inter- and intra-observer variability. 
 
Ideally, a measurement method eliminates these variabilities, is as fast as the linear dimensions method, and is as precise as the manual delineation method. Increasing consistency in measurement enables improved differentiation between actual growth and measurement errors. Additionally, given that in the medical records  of many of the MRI scans there are incomplete or missing linear dimension measurements of the tumor, fast volume measurement would enable systematic studies with a large amount of tumor volume data. In particular, to determine the best fitting growth function (see e.g.,~\cite{heesterman2019mathematical}), increasing the number of included patients and the number of measurements per patient would add to the quality of the results. Additionally, recently it was concluded that studies with long-term growth monitoring are needed to evaluate the risks of symptom development and treatment outcomes related to tumor growth rates~\cite{tooker2023natural}. 
 

Here, we propose an automatic segmentation model trained on 3D Time-Of-Flight (TOF) gadolinium-enhanced MRI scans and corresponding manual volumetric paraganglioma delineations using deep learning (nnU-Net~\cite{isensee2018nnu,isensee2021nnu,siddique2021u}). The resulting model is evaluated both qualitatively and quantitatively, which includes an assessment of variation that is compared to the inter-observer variability in manual volumetric paraganglioma delineation. Finally, using this model to automatically obtain tumor volume measurements for a large data set (with in total 311 tumors), and linking the tumors over time using registration, we showcase how these additional volume measurements affect the fit of known growth functions.

\section{MATERIALS AND METHODS}

\subsection{Data}\label{sec:data}
For creating the segmentation model, we use retrospective data from 93 patients with paraganglioma from Leiden University Medical Center. 16 of these 93 patients were selected because they received radiotherapy and thus tumor delineations were available, which were based on CT scans. These scans were acquired between 01-01-2012 and 31-12-2021. The remaining 77 patients were selected such that at least two MRI scans were acquired between 01-01-2000 and 31-12-2020, and such that, when combined with the 16 patients, the distribution of the location of the tumors of the 93 patients matches the distribution in location of tumors of all paraganglioma patients at LUMC (based on radiology reports). We include one 3D TOF gadolinium-enhanced MRI scan per patient.
Research shows that 3D TOF gadolinium-enhanced MRI scans are superior to 1-weighted, dual T2-weighted, and fat-suppressed MRI scans when it comes to detecting paraganglioma~\cite{van2004head}. These scans were performed on an 1.5T or a 3T Philips Medical Systems scanner. Further details on these MRI scans can be found in \autoref{tab:mr_details}. These scans were split into a cross-validation set (73 cases) and an independent test set (20 cases).

\begin{table}[h]
\centering 
\caption{Details on the MRI scans.}
\scalebox{0.7}{
\begin{tabular}{ | c | c | c | c |} 
\hline 
\textbf{Technical feature} & median & min & max \\ 
\hline 
\textbf{in-plane shape} & 512 \texttimes \ 512\  & 256 \texttimes \ 256\  & 768 \texttimes \ 768\  \\ 
\hline
\textbf{\#slices} & 200 & 150 & 304 \\ 
\hline
\textbf{in-plane resolution ($mm$)} & 0.39 \texttimes \ 0.39\  & 0.32 \texttimes \ 0.32\  & 1.02 \texttimes \ 1.02\  \\ 
\hline
\textbf{slice tickness ($mm$)} & 0.75 & 0.5 & 0.8 \\ 
\hline
\textbf{TE i.e. echo time ($msec$)}& 20.0 & 16.0 & 22.0 \\ 
\hline
\textbf{TR i.e. repetition time($msec$)} & 3.45 & 2.3 & 6.9 \\ 
\hline 
\end{tabular}}
\label{tab:mr_details}
\end{table}

\begin{figure}
  \centering
\includegraphics[width=\textwidth]{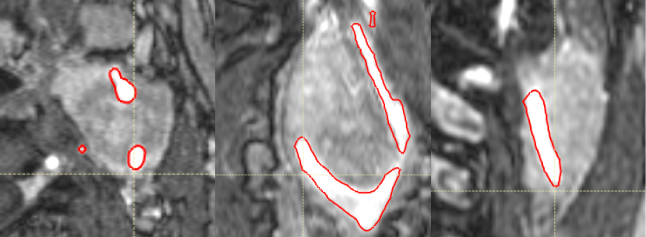}
    \caption{Example of delineation of the vessels. The left image shows the transversal view, the middle image shows the coronal view and the right image shows the sagittal view.  }
    \label{fig:vessels}
\end{figure}

The delineations for the 16 patients that received radiotherapy were transferred to the corresponding MRI scans by affine registration with Elastix software using default settings~\cite{klein2009elastix}. After registration, the registered delineations were visually inspected and possibly adjusted by a researcher (E.M.C.S) under the supervision of a senior radiation oncologist (M.R.). For the remaining 77 patients, one MRI scan was randomly selected from the set of available scans, and manual delineations were created by a researcher (E.M.C.S) under the supervision of a senior otorhinolaryngologist (J.C.J.). Every tumor (total of 147, see \autoref{tab:tumor_details}) was delineated using RayStation 10B Research Software (v10.1.1) as a separate Region-of-Interest (ROI). Further, a separate delineation was made for the major blood vessels surrounding or inside the tumor with the aim of their exclusion in the automatically generated delineations. This was done by subtracting the vessel from the final delineation of the tumor. An example of the vessels delineation is shown in \autoref{fig:vessels}.

\begin{table}[htbp]
\centering 
\caption{Details on number of tumors per location and the volumes of these tumors.}
\scalebox{0.7}{
\begin{tabular}{ | c | c | c | c | c | c | } 
\hline 
\textbf{Location (number)} & Carotid (n=69) & Vagal (n=30) & Jugulotympanic (n=22) & Multiple loc. (n=26)  \\ 
\hline 
\textbf{Volume ($cc$): median (min - max)}& 2.6 (0.14 - 50.72) & 2.87 (0.33-81.79) & 10.13 (0.82 - 88.88) & 55.53 (7.22 - 1329.65)  \\ 
\hline 
\end{tabular}}
\label{tab:tumor_details}
\end{table}

For the observer variation study, we included 10 patients, different from the patients whose scans were used to train and test the model. We randomly selected from each patient one MRI scan from the set of available scans. In total 19 tumors were visible in these scans. Furthermore, to fit growth functions, we used all available 3D TOF gadolinium-enhanced MRI scans of 208 paraganglioma patients (this includes patients whose scans were included for the model training/testing and the observer study). These 208 patients were the ones for whom the model detected the same tumor in at least 3 scans.

\subsection{ML-pipeline}
To use nnU-Net, we first need to convert the ROI labels to segmentation masks. The best approach to do so, we argue, is to give each tumor the same label. If each tumor would be labeled separately, there needs to be some (geometric) logic in labeling for the model to learn which tumor has which label. However, this would complicate training since the paraganglioma can spread over multiple locations. Assume for instance a reference  ROI of a ``multiple locations'' volume instead of one carotid volume and a vagal volume. It might be challenging to determine whether these two volumes are separate or actually attached. Now assume that the segmentation model outputs a mask with one carotid volume and one vagal volume close to each other. The model's performance on that scan would be the lowest it can be, whereas the result might be appropriate. Therefore, we give the same label to every tumor. 

nn-UNet automatically performs training with both 2D and 3D architectures. Using a 3D architecture was found to be preferable. Furthermore, we use 5-fold cross-validation, as is standard in nn-UNet. nn-UNet automatically performs pre-processing, and for our data a patch size of 80x192x160 voxels was selected.
nn-UNet has some built-in functionality for post-processing, such as only keeping the biggest connected component for each
label. Since the label for all of our tumors is the same, post-processing the masks this way is not an option. Instead, we use a
different procedure.  First, we compute connected components for each mask. Then, we calculate
the volume of each connected component and discard it if its volume is smaller than 0.1 $cc$. We deem this threshold safe, since the smallest tumor in our data set is 0.14 $cc$, and most tumors are
much bigger. If the volume is bigger than 0.1 $cc$, we calculate the centroid of the volume. We look up the voxel at this centroid
in the reference and link the tumor in the prediction mask to the tumor at that place (if any). In looking up this voxel, we use the reference without the vessels
subtracted. Otherwise, if the centroid is at the vessel, this would hamper the linking. Once the volumes in the reference are
linked to the volumes in the prediction mask, the per-tumor metrics can be calculated.

\subsection{Evaluation}
\noindent\underline{\textit{Qualitative evaluation}} Visual evaluation of the test set (containing 32 tumors) is done to determine whether the model's output is (at least) equal to human performance from a qualitative perspective. 
A senior otorhinolaryngologist (J.C.J.) rates for every tumor which delineation is better (automatic or manual) or whether they are equally good, without knowing which is automatic and which is manual. They also rate whether a delineation is appropriate to use. 
Furthermore, all delineations produced by the model on the cross-validation set are visually inspected to assess whether they are appropriate to use, as rated by a researcher (E.M.C.S).  Additionally, we inspect possible additional and missing delinations to report precision and recall over all tumors. 

\noindent\underline{\textit{Quantitative evaluation}}
We calculate the Dice score, 95\% percentile of the Hausdorff distance, average surface distance, and relative volume error for each tumor. We compute the average surface distance by first taking for every voxel of the contour of the reference mask the distance to the closest voxel in the contour of the prediction mask, and vice versa. Then, we take the average of these distances.
Note that this is symmetrical: the average surface distance between the reference and prediction is the same as the average surface distance between the prediction and the reference. We report combined cross-validation scores over the validation folds (one model) and test scores over the resulting ensemble of models. To not skew results to measurements that would not be used anyway, we report the number of tumors deemed inappropriate during visual inspection and leave them out when computing quantitative metrics. 

\noindent\underline{\textit{Observer variation study}}
We compare the observer variation between multiple observers with the variation between the model and the observers (in terms of Dice score and relative difference in volume). For this purpose, additional delineations are made for 10 scans by three observers (E.M.C.S., M.R., J.C.J.). 
To ensure that all observers delineate the same set of tumors, they are given the MRI scans as well as a description regarding the number and location of the tumors to delineate per scan. The observers are instructed to be as precise as possible and to not include major blood vessels that surround or run through the tumor. 
We test whether the variation between two human annotators is from the same distribution as the variation between one of those human annotators and the model using the Wilcoxon signed rank test with Bonferroni-Holm correction using a p-value of 0.05.


\subsection{Growth-curves study}
To evaluate the potential of utilizing this segmentation model for modeling tumor growth, we do a similar analysis as in~\cite{heesterman2019mathematical}. We perform this analysis with the hypothesis that when using data of more patients and more volume measurements per patient, we can improve the quality of the results. We run the segmentation model on all scans of 208 paraganglioma patients. We then perform affine registration with Elastix software using standard settings to each pair of subsequent scans and link the tumors between the subsequent scans. Since we are now merely linking the tumors over time and not evaluating the model's output, we can use the Dice score to link tumors since we do not use it as an evaluation metric on this data. We then visually inspect the volume over time of these linked tumors. We specifically look for anomalies that could point to problematic linking or volume measurement inaccuracies. Since the tumors rarely decrease in volume (beyond a possible measurement error) and are primarily slow growing~\cite{michalowska2017growth,langerman2012natural,carlson2015natural,prasad2014role}, this mainly includes checking for big increases or decreases in volume w.r.t. the last measurement. If an anomaly is detected, we either verify it by checking if similar patterns are described in the radiology report, correct the problem if possible, or omit it from the data set. Furthermore, we omit volume measurements of tumors after treatment (surgery or radiotherapy). We then fit seven known growth curves (Linear, Exponential, Mendelsohn, Gompertz, Logistic, Spratt, and Bertalanffy) to each tumor under the constraints based on domain knowledge that the volume at birth should be below 0.01 $cc$, and the volume at the age of 100 can not be bigger than 1500 $cc$. To perform this constrained fit, we employ real-valued optimization using RV-GOMEA~\cite{bouter2017exploiting}, using default settings and a budget of 90 seconds per optimisation run. We compare the results of previous research~\cite{heesterman2019mathematical} (3 volume measurements per tumor), to our method using 3 volume measurements per tumor, and to our method using all available data points per tumor (median: 5, range: 3-15 volume measurements per tumor) in terms of the Root Mean Square Error (RMSE) between the fitted curve and the volume measurements. The results of previous research are based on a data set with 77 tumors from 44 patients using the linear dimensions measurement method~\cite{heesterman2019mathematical} (with an overlap of 39 patients with our data set). 

\section{RESULTS}

\begin{figure}
    \centering
\includegraphics[width=\textwidth]{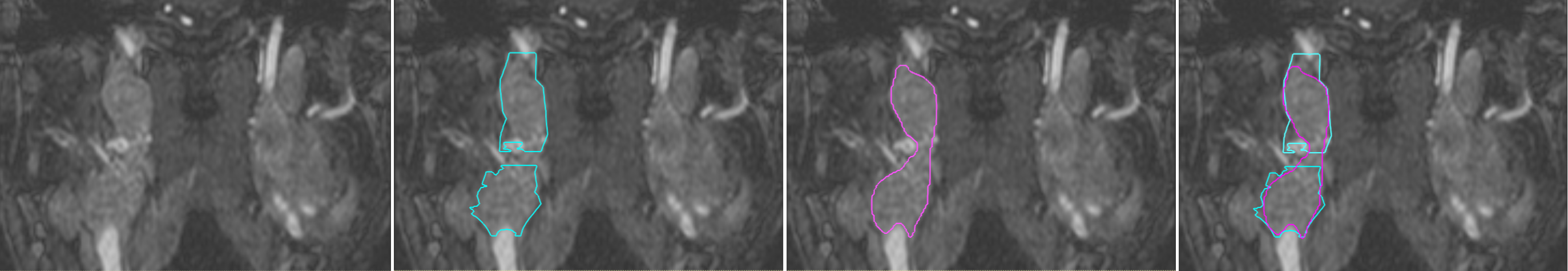}
    \caption{Example of tumor detected over multiple locations by the model with two separate delineations in the reference (coronal view, blue: human/reference, pink: model). It shows an example of when the model delineates the tumors as one tumor while the human/reference delineation consists of two tumors.} 
    \label{fig:attached}
\end{figure}

\subsection{Qualitative evaluation results}
\underline{\textit{Cross-validation set}} In the cross-validation set, the reference scans contained a total of 115 tumors. Of the 115 reference tumors, there were 10 tumors that were not linked initially to tumors found by the model. In some of these cases, the tumors were detected but as a connected tumor over multiple locations instead of two separate tumors and thus could not be linked by the centroids (6 cases, for example as in \autoref{fig:attached}). In some cases it might be hard to determine whether the tumors are separate, and if so how to separate them. Therefore, both strategies might be reasonable. In other cases, the detected volume included a vessel such that the centroid was below the actual tumor volume in the reference and it could thus not be linked (1 case), or the tumor was detected but not fully such that the volume was \textless 0.1 $cc$ (2 cases), or it was just not detected (1 case). We considered the partially correctly detected and not detected tumors as undetected tumors and thus deemed 4 of all tumors undetected after visual inspection. This results in a recall of 97 \%. 

In addition to the reference 115 tumors, 23 volumes were detected by the model. In some cases, these were highly likely to be rightfully detected and thus probably falsely missing in the reference (7 cases). For one of those cases, there already was a suspicion of a tumor, but it was only confirmed two radiology reports later. For another one of these cases, there had been a suspicion since four years prior to the acquired MRI scan, but could not be confirmed nor invalidated as this was the last radiology report available. For the other 5 of those cases the tumor was described in the radiology report, but not in the reference of the manual delineations. Because of cases such as in \autoref{fig:attached}, it might be difficult to assess from the radiology report how many tumors are present in the scan and thus smaller tumors can easily be missed when delineating. In the other cases there were other issues. In particular, it either concerned a tumor residual due to surgery (2 cases), or the tumors were in the reference but not linked due to the same reason the reference could not be linked to the prediction mask (3 cases), or they were wrongly detected (11 cases of which 5 cases in one scan). We only deem the wrongly detected volumes to be false positives. This results in a precision of 92 \%.

In total, we judged the output of the model unusable for 9 tumors (5 jugulotympanic tumors, 3 vagal tumors, and 1 tumor spreading over multiple locations). 4 of those tumors were undetected or only partly correctly detected (3 vagal tumors and 1 jugulotympanic). The remaining 5 tumors can be considered to be challenging cases to delineate, even for humans. Of the remaining tumors, 4 are jugulotympanic tumors, which are tumors that can be hard to detect and to delineate, specifically because they can be intertwined or confused with the jugular bulb. The observer study confirms that this is difficult for humans too. The multiple locations one is an extraordinary case of an unusually big paraganglioma. It was clear that the model had trouble segmenting this tumor since it found 5 additional volumes in this scan. Furthermore, the reference for one carotid volume was judged uncertain during the visual inspection because part of the delineation could have been mistakenly included (it might actually be a vessel). We leave this tumor out of further analysis. The final results of visual inspection as used for the quantitative study are
summarized in \autoref{tab:detection_crossval}.

\begin{table}[]
\centering 

\caption{Summary of qualitative analysis  for the cross validation set (as used for the quantitative study). 3 tumors were detected as tumors over multiple locations (denoted as +3) by the model, while they were delineated in the reference as either one carotid and one vagal tumor or one carotid and one jugulotympanicum tumor (denoted as -3,-1 and -2).  }
\scalebox{0.7}{
\begin{tabular}{l|l|l|l|l|l|}
\cline{2-6}
                                                                      & Carotid          & Vagal            & Jugulotympanicum & Multiple loc.    & Uncategorised \\ \hline
\multicolumn{1}{|l|}{\textbf{Total in reference}}          & \textbf{56}      & \textbf{21}      & \textbf{16}      & \textbf{22}      & \textbf{-}    \\ \hline
\multicolumn{1}{|l|}{\textbf{Correctly detected by model}}            & \textbf{55 (-3)} & \textbf{18 (-1)} & \textbf{11 (-2)} & \textbf{21 (+3)} & \textbf{-}    \\ \hline
\multicolumn{1}{|l|}{Not (or partly) detected}                        & -                & 3                & 1                & -                & -             \\ \hline
\multicolumn{1}{|l|}{Inappropriate tumor delineation from model}      & -                & -                & 4                & 1                & -             \\ \hline
\multicolumn{1}{|l|}{Falsely detected}                                & -                & -                & -                & -                & 11            \\ \hline
\multicolumn{1}{|l|}{Missing/inappropriate reference, tumor residual} & 1                & -                & -                & -                & 9             \\ \hline
\end{tabular}}
\label{tab:detection_crossval}
\end{table}

\begin{figure}
   \centering
\includegraphics[width=0.8\textwidth]{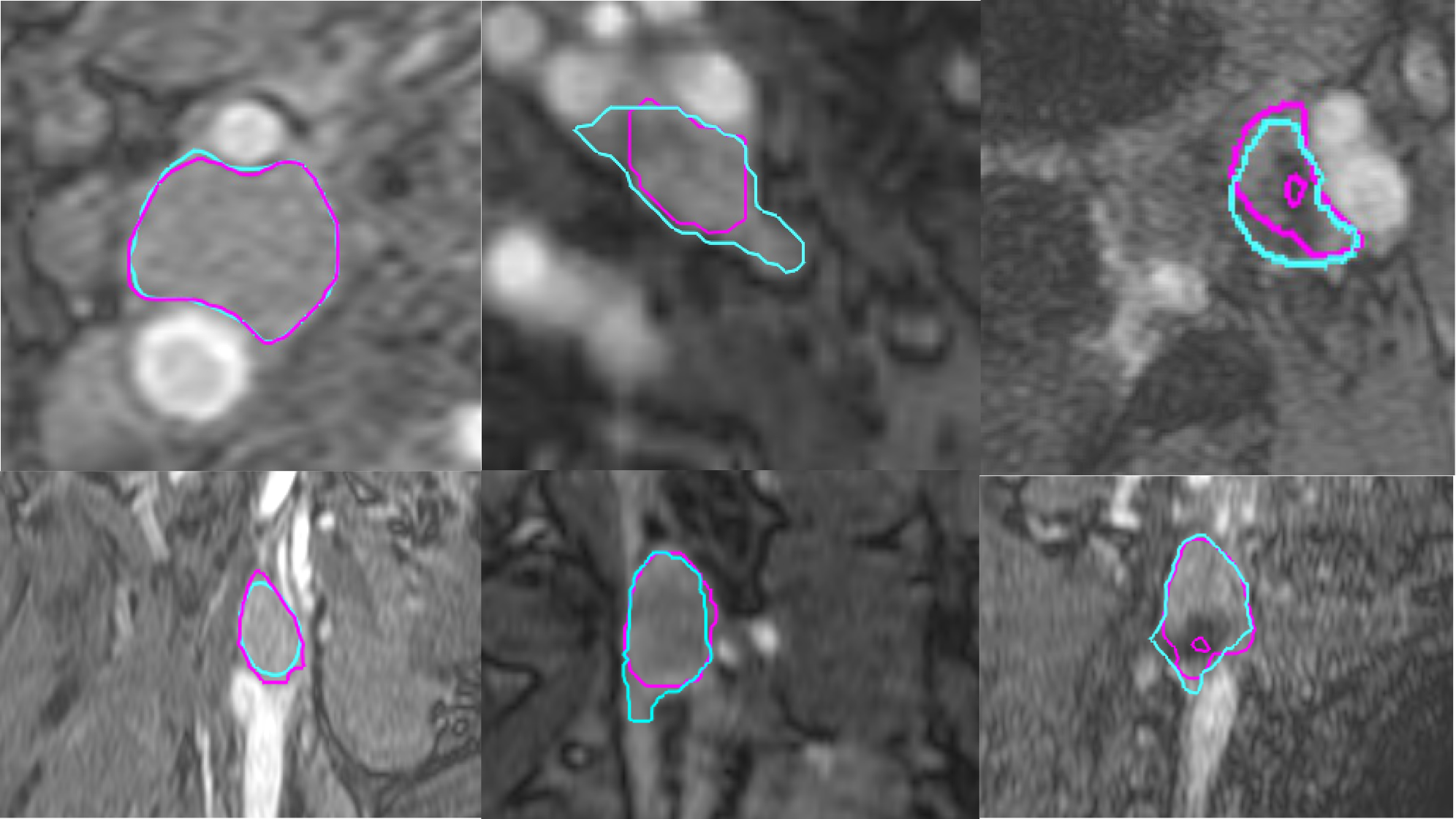}
    \caption{Examples of patients shown during the blinded study (blue: human, pink: model). In the first column, the delineations were judged equal. In the second column, the automatic delineation (pink) was preferred. In this case it seems as if a vessel was wrongly included in the human delineation. In the third column, the human (blue) delineation was preferred. In this case it seems as if the model tries to exclude the black hollow space. }
    \label{fig:visualstudy}
\end{figure}
\noindent \underline{\textit{Test set}}
In the test set, 3 volumes were wrongly marked as tumors. This results in a 91 \% precision. 
All tumors were detected, but one carotid tumor was not fully detected such that the volume was \textless 0.1 $cc$. This results in a recall of 97 \%.

In the blinded study, we included all the tumors in the test set except for the one partially undetected. In this study, in 11/31 cases, the automatic delineation was rated better, in 10/31 cases the delineations were judged equal, and in 10/31 cases, the manual delineation was rated better. \autoref{fig:visualstudy}, shows an example for each of these three cases. During the blinded study, one of the automatic delineations was judged not appropriate, since it only included the tumor partially. The final results of visual inspection as used for the quantitative study are summarized in \autoref{tab:detection_test}. 

\begin{table}[]
\centering 

\caption{Summary of qualitative analysis of the test set (as used for the quantitative study). }
\scalebox{0.7}{
\begin{tabular}{l|l|l|l|l|l|}
\cline{2-6}
                                                                      & Carotid          & Vagal            & Jugulotympanicum & Multiple loc.    & Uncategorised \\ \hline
\multicolumn{1}{|l|}{\textbf{Total in reference}}          & \textbf{13}      & \textbf{9}      & \textbf{6}      & \textbf{4}      & \textbf{-}    \\ \hline
\multicolumn{1}{|l|}{\textbf{Correctly detected by model}}            & \textbf{12} & \textbf{8} & \textbf{6} & \textbf{4} & \textbf{-}    \\ \hline
\multicolumn{1}{|l|}{Not (or partly) detected}                        & 1                & -                & -               & -                & -             \\ \hline
\multicolumn{1}{|l|}{Inappropriate tumor delineation from model}      & -                & 1               & -               & -                & -             \\ \hline
\multicolumn{1}{|l|}{Falsely detected}                                & -                & -                & -                & -                & 3            \\ \hline

\end{tabular}}
\label{tab:detection_test}
\end{table}

\begin{table}
\centering 
\caption{Quantitative scores for cross-validation and test set (mean $\pm$ SD).}
\label{tab:val_scores}
\scalebox{0.7}{
\def\arraystretch{0.9}
\begin{tabular}{|c | c | c | c | c | c | c | c | } 
\hline 
Set & Metric & Carotid \textgreater 1 $cc$ & Carotid \textless 1 $cc$ & Vagal \textgreater 1 $cc$ & Vagal \textless 1 $cc$ & Jugulotympanic   & Multiple loc.  \\ 
\hline
Validation& Count & 38 & 14 & 14 & 3 & 9 & 24 \\ 
 & Dice score & 0.88 $\pm$ 0.05 & 0.78 $\pm$ 0.10 & 0.88 $\pm$ 0.05 & 0.77 $\pm$ 0.15 & 0.86 $\pm$ 0.05 & 0.88 $\pm$ 0.05  \\ 
 & 95 \% Hausdorff (mm) & 2.64 $\pm$ 1.42 & 1.60 $\pm$ 0.56 & 4.06 $\pm$ 4.37 & 2.52 $\pm$ 1.80 & 4.27 $\pm$ 3.04 & 4.46 $\pm$ 2.23  \\ 
 & Avg. surface distance (mm) & 0.65 $\pm$ 0.28 & 0.50 $\pm$ 0.25 & 0.84 $\pm$ 0.65 & 0.61 $\pm$ 0.46 & 1.16 $\pm$ 0.78 & 1.21 $\pm$ 0.65  \\ 
 & Volume diff. (\%) & 12.02 $\pm$ 10.10 & 29.76 $\pm$ 52.05 & 11.30 $\pm$ 10.69 & 48.18 $\pm$ 67.60 & 12.59 $\pm$ 15.23 & 13.60 $\pm$ 13.85  \\
\hline
Test& Count  & 9 & 3 & 6 & 2 & 6 & 4 \\ 

 & Dice score & 0.90 $\pm$ 0.02 & 0.82 $\pm$ 0.12 & 0.91 $\pm$ 0.03 & 0.87 $\pm$ 0.02 & 0.80 $\pm$ 0.07 & 0.89 $\pm$ 0.05  \\ 
 & 95 \% Hausdorff (mm) & 1.90 $\pm$ 0.71 & 1.12 $\pm$ 0.61 & 4.22 $\pm$ 6.01 & 0.86 $\pm$ 0.02 & 4.45 $\pm$ 1.86 & 2.37 $\pm$ 0.52  \\ 
 & Avg. surface distance (mm) & 0.48 $\pm$ 0.14 & 0.28 $\pm$ 0.18 & 0.75 $\pm$ 0.70 & 0.26 $\pm$ 0.00 & 1.19 $\pm$ 0.52 & 0.68 $\pm$ 0.20  \\ 
 & Volume diff. (\%)  & 8.39 $\pm$ 5.85 & 25.32 $\pm$ 27.65 & 8.89 $\pm$ 6.82 & 1.94 $\pm$ 1.09 & 27.10 $\pm$ 30.40 & 12.24 $\pm$ 9.73  \\ 
\hline
\end{tabular}}
\end{table}

\subsection{Quantitative evaluation results}
In \autoref{tab:val_scores}, we report the quantitative scores for the cross-validation set as well as the test set. Notice that the numbers reported here include only the tumors that were judged appropriate during the visual study. The scores would be worse if all detected tumors were included, but it would give a skewed view of the performance of the model over the appropriate tumor delineations. 

Additionally, we separately report cross-validation and test scores for tumors \textgreater 1 $cc$ and \textless 1 $cc$, since the difference in the volume of the tumors \textless 1 $cc$ is more unstable (i.e., including an extra slice in the delineation can lead to a significant change in volume percentage). From these scores we conclude that, generally, the model does well in this task. The model performs slightly better in terms of volume difference and Dice score for the multiple location tumors, bigger carotid, and bigger vagal tumors than for the smaller carotid, smaller vagal, and jugulotympanic tumors. In terms of 95 $\%$ Hausdorff distance, the model performs slightly better on the carotid tumors (considering that generally, a smaller volume would yield smaller Hausdorff distances, given that the delineation is useful). The mean average surface distance is below 1 mm for all locations except for the jugulotympanic location. This can be explained by the fact that the jugulotympanic tumor is generally more difficult to delineate.

\begin{figure}
\centering
\includegraphics[width=\textwidth]{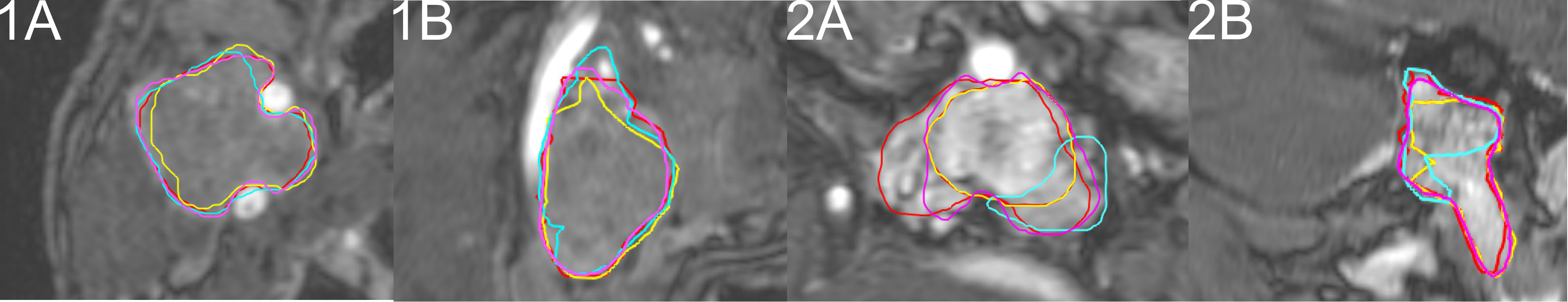}
    \caption{Examples of different delineations of the different observers (pink = model, red = h.a.1, yellow = h.a.2 and blue = h.a.3.), with A in the transversal plane and B in the sagittal plane. The delineations of the first tumor (carotid tumor) are more similar than the delineations of the second tumor (jugulotympanic). }
    \label{fig:observerstudy}
\end{figure}

\begin{table}
\caption{Variation between multiple human annotators (h.a.), and between the model and these human annotators (mean $\pm$ SD). Volume difference is normalized using the mean volume of the two observers (i.e., a combination of two human annotators or one human annotator and the model).}
\label{tab:obs-1}
\centering
\scalebox{0.8}{
\def\arraystretch{0.9}
\begin{tabular}{ | c | c | c | c | c | c | c | } 
\hline 
Metric & First observer & Second observer & Carotid & Vagal & Jugulotympanic & Multiple loc. \\ 
\hline 
\multirow{6}{*}{Dice score}
 & h.a. 1 & h.a. 2 & 0.84 $\pm$ 0.03 & 0.89 $\pm$ 0.06 & 0.71 $\pm$ 0.06 & 0.89 $\pm$ 0.03  \\ 
 & h.a. 2 & h.a. 3 & 0.78 $\pm$ 0.15 & 0.89 $\pm$ 0.04 & 0.67 $\pm$ 0.12 & 0.88 $\pm$ 0.02  \\ 
& h.a. 1 & h.a. 3 & 0.78 $\pm$ 0.14 & 0.89 $\pm$ 0.04 & 0.66 $\pm$ 0.11 & 0.88 $\pm$ 0.03  \\ \cline{2-7}
 & h.a. 1 & model & 0.86 $\pm$ 0.04 & 0.90 $\pm$ 0.07 & 0.79 $\pm$ 0.10 & 0.90 $\pm$ 0.02  \\ 
& h.a. 2 & model & 0.86 $\pm$ 0.03 & 0.92 $\pm$ 0.03 & 0.69 $\pm$ 0.20 & 0.91 $\pm$ 0.03  \\ 
 & h.a. 3 & model & 0.81 $\pm$ 0.12 & 0.89 $\pm$ 0.06 & 0.62 $\pm$ 0.13 & 0.89 $\pm$ 0.02  \\ 
\hline
\multirow{6}{*}{Volume diff. (\%)}
 & h.a. 1 & h.a. 2 & 8.23 $\pm$ 5.18 & 2.89 $\pm$ 0.64 & 36.76 $\pm$ 18.39 & 4.86 $\pm$ 1.42  \\ 
 & h.a. 2 & h.a. 3 & 29.40 $\pm$ 33.51 & 6.33 $\pm$ 5.00 & 24.96 $\pm$ 16.67 & 6.66 $\pm$ 5.62  \\ 
 & h.a. 1 & h.a. 3 & 27.09 $\pm$ 32.81 & 7.05 $\pm$ 7.11 & 45.93 $\pm$ 22.70 & 6.45 $\pm$ 4.46  \\ \cline{2-7}
& h.a. 1 & model & 7.70 $\pm$ 7.85 & 2.14 $\pm$ 2.66 & 25.36 $\pm$ 27.60 & 2.70 $\pm$ 1.70  \\  
 & h.a. 2 & model & 10.77 $\pm$ 8.55 & 4.48 $\pm$ 3.65 & 55.76 $\pm$ 39.02 & 5.54 $\pm$ 3.50  \\ 
 & h.a. 3 & model & 23.90 $\pm$ 25.68 & 8.63 $\pm$ 4.27 & 61.20 $\pm$ 26.06 & 6.31 $\pm$ 4.70  \\ 
\hline
\end{tabular}}
\end{table} 
\subsection{Observer variation study results}
In the observer variation study, again, we first performed a visual inspection before computing quantitative metrics. 
For two jugolotympanic tumors, the manual delineations of all three human annotators were not overlapping. We left these out of further analysis and evaluated the model on the 16 remaining tumors.
Two volumes were wrongly detected by the model, and one jugulotympanic tumor was not detected. 
In this set, the precision was 89 $\%$ and the recall was 94 $\%$.
\autoref{tab:obs-1} gives an overview of the inter-observer variation in manual delineation and the variation between the model and the human annotators. \autoref{fig:observerstudy} shows examples of delineations of the model and the three human annotators. There is no statistical significant difference in Dice score or volume difference between two human annotators, and between a human annotator and the model for 11 out of 12 tests. In 1 out of 12 tests the Dice score between one observer and the model was significantly higher than the Dice score between two observers, denoting even less variation between the model and the human annotator than among human annotators.    

\begin{figure}
  \centering
\includegraphics[width=\textwidth]{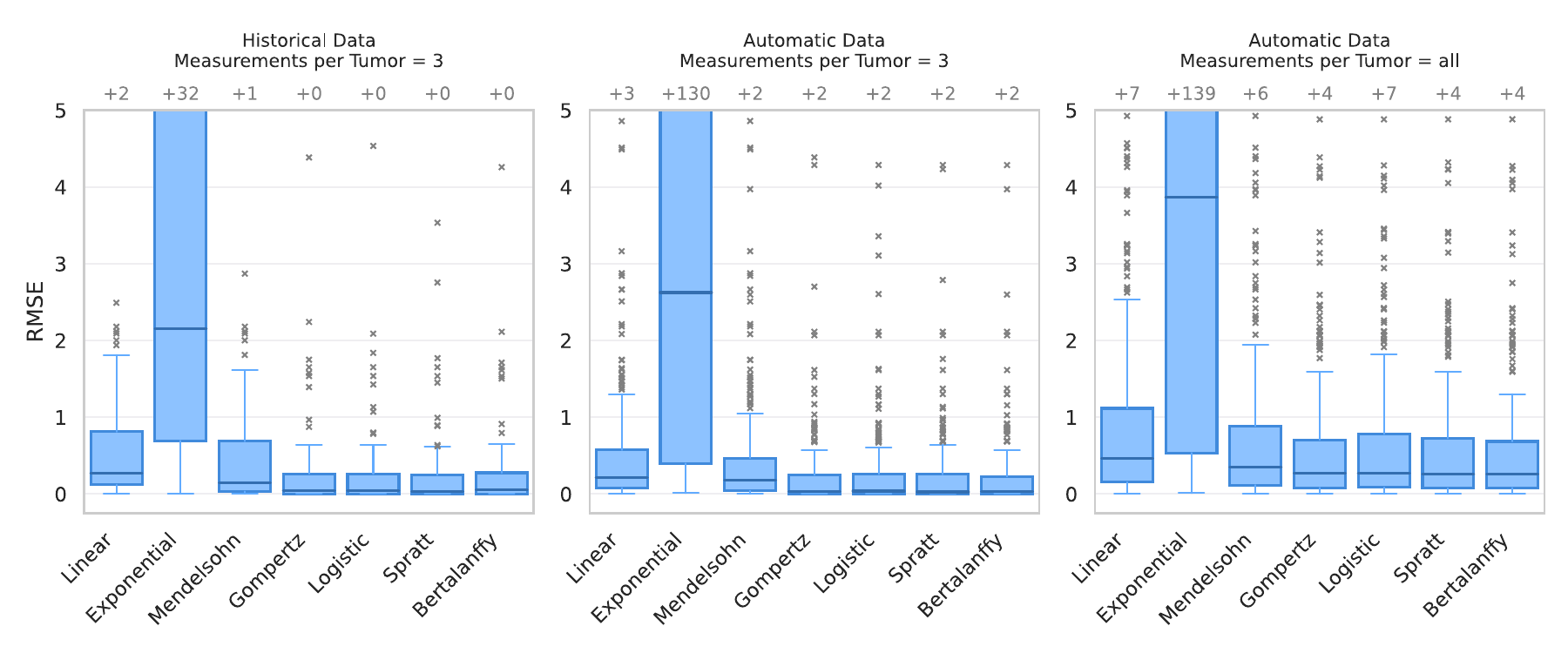}
    \caption{Box plots of the RMSE (in cc) of fitted functions on the three data sets. Numbers above the boxplots indicate the number of tumors for which RMSE $> 5$.}
    \label{fig:boxplots}
\end{figure}

\subsection{Growth-curve results}
To fit the growth curves, we used data of 311 tumors in 208 patients with a median of 5 (range: 3-15) volume measurements per tumor. 
\autoref{fig:boxplots} shows results of fitting the set of previously investigated growth curves to the volume measurements. The box plot for the automatic method (limited to 3 data points) shows similar patterns as the box plot for the historical data set (limited to 3 data points), i.e., the s-shaped (Gompertz, Logistic, Spratt and Bertalanffy) functions appear to give a better fit than the ever-growing functions (Linear, Exponential and Mendelsohn). However, when using all available points per tumor, this difference (specifically between the Linear and Mendelsohn functions) becomes smaller. Considering that the s-shaped functions can be approximately linear at a specific interval, one could argue, using Occam's razor, that the linear one would be the preferred function when the fits are approximately equal over the interval of these volume measurements. Additionally, the s-shaped functions more often find a fit of a volume above 1000 $cc$ at the age of 100, including many times the absolute maximum of 1500 $cc$. This seems implausible given that in the clinic, it is exceptional for a tumor to be above 1000 $cc$. On the one hand, the fits of the linear function do not have this implausible property, but on the other hand the linear function can not explain that some paraganglioma tumors appear to stay at a stable volume (after a certain age). 
Whereas before it was concluded that the s-shaped functions, and specifically the Bertalanffy function, was preferred, we now conclude that, based on these outcomes, it is uncertain whether one of these functions is always the preferred one. This highlights the need for more research. Perhaps another function not explored here fits better, or, for different patients, different growth functions fit better, e.g., due to different gene mutations.

\begin{figure}[t]
   \centering
\includegraphics[width=0.8\textwidth]{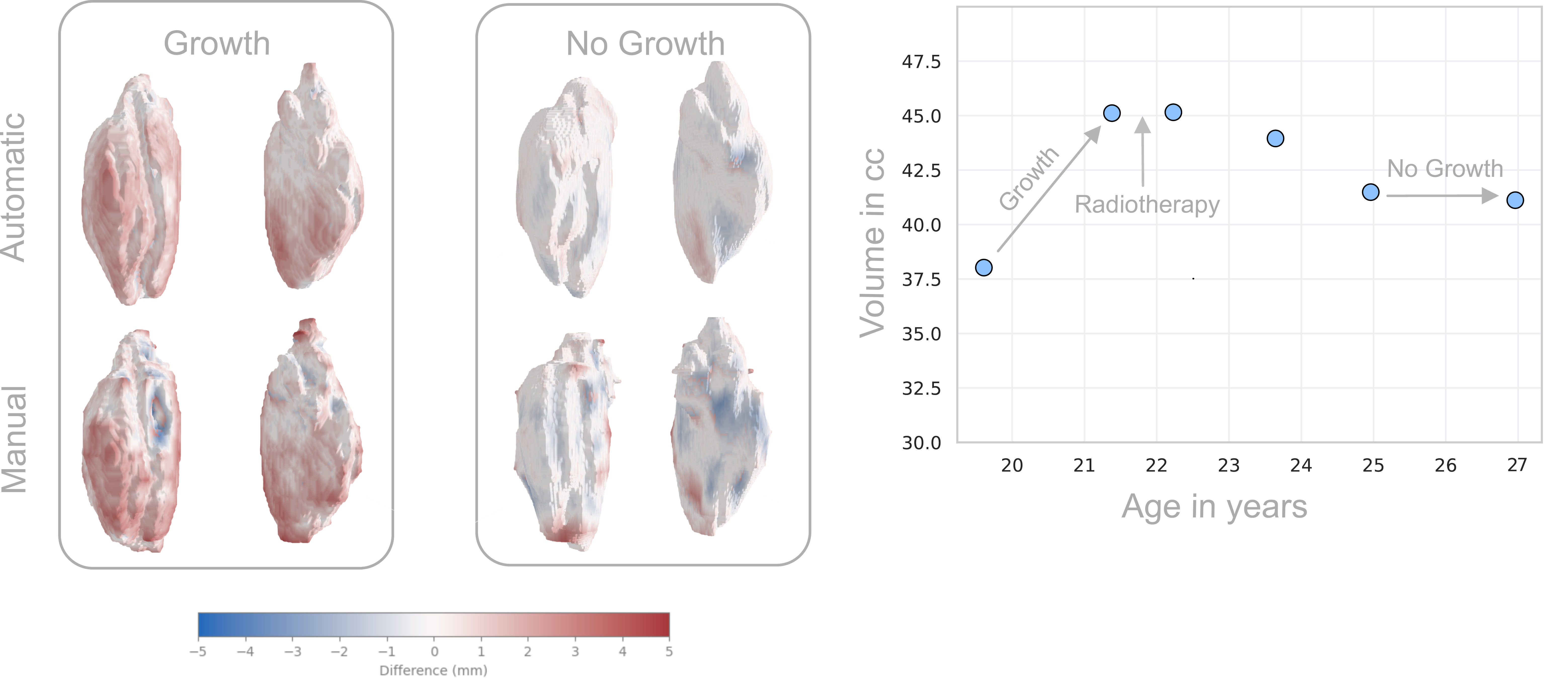}
    \caption{On the left, we show a 3D visualization of tumor volume differences between two subsequent MRI scans obtained via manual delineation versus auto-segmentation. The shown 3D volume is that of the last of the two scans. Red indicates growth, blue indicates decline, and white indicates no difference compared to the first scan (more intense colors represent larger differences). Two viewpoints of the same volume are illustrated. On the right, we show the volume in cc of the visualized tumor (based on auto-segmentation) against the age of the patient in years.}
    \label{fig:3D}
\end{figure}
\section{DISCUSSION AND CONCLUSION}
We present, for the first time, a U-Net model trained specifically for paraganglioma tumors in the head and neck area and performed several quantitative and qualitative evaluation studies to demonstrate the model's capabilities. Although some of the tumors are falsely detected or undetected by the model, the model detects most tumors correctly. Our quantitative results show that the model performs well for most tumors. The fact that the models does not perform well on some tumors can be explained by the fact that some paraganglioma can be hard to delineate, as was confirmed in our observer study. The boundaries of some tumors can be difficult to accurately identify, especially for larger tumors that often diverge more from the typical ellipsoid shape.
Separating tumors from blood vessels can also be difficult, especially for tumors at the jugulotympanic location, because paraganglioma tumors are often intertwined with blood vessels. In this paper, we attempted to eliminate the vessels as much as possible from the tumor delineation reference by introducing a separate delineation of these vessels. Despite this, it might still be hard to tell them apart.

We performed a visual evaluation study of the human delineations and the model's delineations of the tumors in the test set. Although there are differences between the human delineation and model's delineation, the model's delineations were preferred approximately the same number of times as the human delineations. Even for the cases where the human delineation was preferred, the model's delineation were only rejected in one of the cases. Therefore, there may be more than one reasonable way to delineate a tumor, and thus a way to quantify this uncertainty through multiple alternative tumor delineations may be more useful and sensible than aiming for a higher quantitative score.

Factors beyond the performance of the model compared to human performance may also be of interest to growth monitoring. However, perhaps the most important additional factor is the consistency of the measurement. Consistency of measurement enables for a fair comparison of volume measurement: if a part of tissue is included in one measurement the same tissue should be included in the next measurement. This, however, is difficult to assess since there is no straightforward measure to quantify the consistency for delineations of a tumor in two consecutive scans. Nevertheless, when the tumor stays stable, i.e., it does not grow, the theoretical expectation of a perfect measurement method would be that the contours of the segmentations match exactly after rigid registration. Furthermore, we would expect the distance difference of a closest point-to-point comparison to be smooth.  \autoref{fig:3D} shows such a point-to-point comparison of the automatic delineation and manual delineation of a tumor for both the growth and the no-growth scenario. The automatic delineation has smoother distance differences, which could indicate a more consistent measurement. However, further research that includes an analysis with more tumors would be needed to further test this hypothesis.  

We showed the possibility of tracking the tumor and its volume over time using the output of the segmentation model, and how this data can be used to fit known growth functions. Although our results show that we can use the automatically gathered volume data to possibly obtain new insights, we conclude that further research is needed to better understand the true nature of growth of paraganglioma. Despite this, our results showed that it is possible to perform a systematic study leveraging a large amount of automatically segmented tumor volume data. This lays the foundation for further systematic studies that leverage large amounts of paraganglioma volume data, and the further development of techniques to study paraganglioma. 
Beyond research purposes, the model could be useful in clinical practice as a tool to aid radiologist in monitoring tumor growth, or to help the radiation oncologist delineate tumors faster for radiation treatment planning, for example, by providing the automatic delineation as a starting point.


\acknowledgments 
This research was funded by the European Commission within the HORIZON Programme (TRUST AI Project, Contract No.: 952060).

\bibliography{report} 

\begin{thebibliography}{10}

\bibitem{pellitteri2004paragangliomas}
Pellitteri, P.~K., Rinaldo, A., Myssiorek, D., Jackson, C.~G., Bradley, P.~J., Devaney, K.~O., Shaha, A.~R., Netterville, J.~L., Manni, J.~J., and Ferlito, A., ``Paragangliomas of the head and neck,'' {\em Oral Oncology}~{\bf 40}(6),  563--575 (2004).

\bibitem{heesterman2019mathematical}
Heesterman, B.~L., Bokhorst, J.-M., de~Pont, L.~M., Verbist, B.~M., Bayley, J.-P., van~der Mey, A.~G., Corssmit, E.~P., Hes, F.~J., van Benthem, P. P.~G., and Jansen, J.~C., ``Mathematical models for tumor growth and the reduction of overtreatment,'' {\em Journal of Neurological Surgery Part B: Skull Base}~{\bf 80}(01),  072--078 (2019).

\bibitem{heesterman2016measurement}
Heesterman, B., Verbist, B., van~der Mey, A., Bayley, J., Corssmit, E., Hes, F., and Jansen, J., ``{Measurement of head and neck paragangliomas: is volumetric analysis worth the effort? A method comparison study},'' {\em Clinical Otolaryngology}~{\bf 41}(5),  571--578 (2016).

\bibitem{tooker2023natural}
Tooker, E.~L., Wiggins~3rd, R.~H., Espahbodi, M., Naumer, A., Buchmann, L.~O., Greenberg, S.~E., and Patel, N.~S., ``{The Natural History of Observed SDHx-Related Head and Neck Paragangliomas Using Three-Dimensional Volumetric Tumor Analysis},'' {\em Otology \& Neurotology}~{\bf 44}(9),  931--940 (2023).

\bibitem{isensee2018nnu}
Isensee, F., Petersen, J., Klein, A., Zimmerer, D., Jaeger, P.~F., Kohl, S., Wasserthal, J., Koehler, G., Norajitra, T., Wirkert, S., et~al., ``{nnU-Net: Self-adapting Framework for U-Net-Based Medical Image Segmentation},'' {\em arXiv preprint arXiv:1809.10486}  (2018).

\bibitem{isensee2021nnu}
Isensee, F., Jaeger, P.~F., Kohl, S. A.~A., Petersen, J., and Maier-Hein, K.~H., ``{nnU-Net: a self-configuring method for deep learning-based biomedical image segmentation},'' {\em Nature Methods}~{\bf 18},  203--211 (Feb 2021).

\bibitem{siddique2021u}
Siddique, N., Paheding, S., Elkin, C.~P., and Devabhaktuni, V., ``{U-Net and its variants for medical image segmentation: A review of theory and applications},'' {\em IEEE Access}~{\bf 9},  82031--82057 (2021).

\bibitem{van2004head}
van~den Berg, R., Verbist, B.~M., Mertens, B.~J., van~der Mey, A.~G., and van Buchem, M.~A., ``Head and neck paragangliomas: improved tumor detection using contrast-enhanced 3d time-of-flight mr angiography as compared with fat-suppressed mr imaging techniques,'' {\em American journal of neuroradiology}~{\bf 25}(5),  863--870 (2004).

\bibitem{klein2009elastix}
Klein, S., Staring, M., Murphy, K., Viergever, M.~A., and Pluim, J.~P., ``Elastix: a toolbox for intensity-based medical image registration,'' {\em IEEE Transactions on Medical Imaging}~{\bf 29}(1),  196--205 (2009).

\bibitem{michalowska2017growth}
Micha{\l}owska, I., {\'C}wik{\l}a, J., Michalski, W., Wyrwicz, L.~S., Prejbisz, A., Szperl, M., Nie{\'c}, D., Neumann, H.~P., Januszewicz, A., and Peczkowska, M., ``Growth rate of paragangliomas related to germline mutations of the sdhx genes,'' {\em Endocrine Practice}~{\bf 23}(3),  342--352 (2017).

\bibitem{langerman2012natural}
Langerman, A., Athavale, S.~M., Rangarajan, S.~V., Sinard, R.~J., and Netterville, J.~L., ``Natural history of cervical paragangliomas: outcomes of observation of 43 patients,'' {\em Archives of Otolaryngology--Head \& Neck Surgery}~{\bf 138}(4),  341--345 (2012).

\bibitem{carlson2015natural}
Carlson, M.~L., Sweeney, A.~D., Wanna, G.~B., Netterville, J.~L., and Haynes, D.~S., ``Natural history of glomus jugulare: a review of 16 tumors managed with primary observation,'' {\em Otolaryngology--Head and Neck Surgery}~{\bf 152}(1),  98--105 (2015).

\bibitem{prasad2014role}
Prasad, S.~C., Mimoune, H.~A., D’Orazio, F., Medina, M., Bacciu, A., Mariani-Costantini, R., Piazza, P., and Sanna, M., ``The role of wait-and-scan and the efficacy of radiotherapy in the treatment of temporal bone paragangliomas,'' {\em Otology \& Neurotology}~{\bf 35}(5),  922--931 (2014).

\bibitem{bouter2017exploiting}
Bouter, A., Alderliesten, T., Witteveen, C., and Bosman, P. A.~N., ``Exploiting linkage information in real-valued optimization with the real-valued gene-pool optimal mixing evolutionary algorithm,'' in [{\em Proceedings of the Genetic and Evolutionary Computation Conference}{\nolinebreak\hspace{0.1em}]},  {\em GECCO '17},  705–712, Association for Computing Machinery, New York, NY, USA (2017).

\end{thebibliography}
\bibliographystyle{spiebib} 

\end{document}